\documentclass[3p,times,twocolumn]{elsarticle}

\usepackage{lineno}
 
\journal{Journal of Internet of Things}

\modulolinenumbers[5] 
\biboptions{numbers,sort&compress}
\usepackage{amsmath,amssymb,amsfonts}
\usepackage{algorithmic}
\usepackage{graphicx}
\usepackage{stfloats}
\usepackage{multirow}
\usepackage{makecell}
\usepackage{latexsym}
\usepackage{textcomp}
\usepackage{booktabs}
\usepackage{amsmath}

\usepackage{xcolor}
\usepackage[colorlinks = true,
            linkcolor = blue,
            urlcolor  = blue,
            citecolor = blue,
            anchorcolor = blue]{hyperref}
\usepackage{subfigure}
\def\BibTeX{{\rm B\kern-.05em{\sc i\kern-.025em b}\kern-.08em
    T\kern-.1667em\lower.7ex\hbox{E}\kern-.125emX}}

\begin{document}

\begin{frontmatter}

\title{Support Vector Machine and YOLO for a Mobile Food Grading System}
  
\author[mymainaddress]{Lili Zhu}
\author[mymainaddress]{Petros Spachos\corref{mycorrespondingauthor}}
\cortext[mycorrespondingauthor]{Corresponding author}
\ead{petros@uoguelph.ca}

\address[mymainaddress]{School of Engineering, University of Guelph, Guelph, Ontario, Canada}

\begin{abstract}
Food quality and safety are of great concern to society since it is an essential guarantee not only for human health but also for social development, and stability. Ensuring food quality and safety is a complex process. All food processing stages should be considered, from cultivating, harvesting and storage to preparation and consumption. Grading is one of the essential processes to control food quality. This paper proposed a mobile visual-based system to evaluate food grading. Specifically, the proposed system acquires images of bananas when they are on moving conveyors. A two-layer image processing system based on machine learning is used to grade bananas, and these two layers are allocated on edge devices and cloud servers, respectively. Support Vector Machine (SVM) is the first layer to classify bananas based on an extracted feature vector composed of color and texture features. Then, the a You Only Look Once (YOLO) v3 model further locating the peel's defected area and determining if the inputs belong to the mid-ripened or well-ripened class. According to experimental results, the first layer's performance achieved an accuracy of 98.5\% while the accuracy of the second layer is 85.7\%, and the overall accuracy is 96.4\%. 
\end{abstract}

\begin{keyword}
Machine Vision\sep Image Processing \sep Food Processing \sep Machine Learning \sep Support Vector Machine.
\end{keyword}

\end{frontmatter} 


\section{Introduction}
Food processing takes raw materials and converts them into suitable forms for modern people's dietary habits.  This process includes a series of physical and chemical changes. During the entire process, the nutrition of the raw materials needs to be maintained to the greatest extent, and the poisonous and harmful substances should be prevented from entering the food. Therefore, food processing is highly valued by food scientists, the food industry, and society~\cite{impactoffoodprocessing}.

The quality inspection of food and agricultural produce is arduous and labor-intensive to meet the increasing expectations and standards of food processing. After years of rapid development, Machine Vision System (MVS) has penetrated several aspects of people's lives. Its high efficiency and accuracy assist various industries to save a large amount of labor~\cite{zhu1, zhu2}. In agriculture, agri-technology and precision farming is an interdisciplinary science that integrates with MVS and utilizes data-intensive methods to achieve high agricultural yields while reducing environmental impact. MVS can acquire image data in a variety of land-based and aerial-based methods and can complete multiple types of tasks as well, such as quality and safety inspection, agriculture produce grading, foreign objects detection, and crop monitoring~\cite{mvforagri}. In food processing, MVS can collect a series of parameters such as size, weight, shape, texture, and color of food, and even many details that human eyes cannot observe. In this way, fatigue and mistakes of workers caused by many repeated labors can be avoided~\cite{mvforfood}. 

Banana is one of the most important tropical fruits and basic staple food for many developing countries. However, banana pests and diseases pose a threat to sustainable production, and banana yellow leaf disease caused by Panama disease is a destructive disease for bananas ~\cite{bananadisease}. Additionally, bananas' ripening process is so rapid that a large number of over-ripened bananas cannot enter the market. As a result, researchers are interested in developing automatic monitoring systems to assist banana management. 
 
In this study, a novel two-layer classifier is proposed, to realize banana grading and defect detection. The classifier comprises the first-layer classifier Support Vector Machine (SVM) and the second-layer classifier YOLOv3. A feature vector containing extracted color and texture information is the input for the first-layer classifier, and the output of the first-layer classifier is connected to the second-layer classifier. This network can provide both the banana ripeness level classification and peel defected area detection to be a solid foundation for further implementing into a visual Internet of Things (IoT) system, such as a smartphone application. Furthermore, a Graphical User Interface (GUI) for a computer application is built for users who need to assess bananas' qualities. Users do not have to enter the warehouse but stay in front of a computer and work on the developed application to evaluate bananas and observe the defects with ease. Finally,the two-layer classifier can be distributed by combining edge computing and cloud computing to improve data communication and computation efficiency.

The rest of this paper is organized as follows: Section~\ref{sec2} reviews the recent related studies. Section~\ref{sec3} introduces the main components of a MVS. Section~\ref{sec4} provides the data set and methodologies used in this research. Section~\ref{sec5} explains and discusses the experiment results, followed by Section~\ref{sec6} which illustrates the proposed Internet of Things application. The conclusions are in the Section~\ref{sec7}.

\section{Related Works} \label{sec2}

MVS applications have been applied to multiple research areas of food processing, such as food safety and quality evaluation, food process monitoring, and foreign object detection. In an MVS, the role of image processing is to guide the operation of the machinery~\cite{imageproforfood}. Regarding food safety and quality evaluation, a multimode tabletop system and adopted spectroscopic technologies for food safety and quality applications is presented in~\cite{method-hyperspectral}. In~\cite{method-fruitdecay}, they designed a hemispherical illumination chamber to illuminate spherical samples and a liquid crystal tunable filter (LCTF) based method that contains two LCTF to acquire images of spherical fruits and segmented hyperspectral fruit images to detect at which maturity stage the sample fruit was. The classification accuracy is 98.5\% for the ``non-rotten" class and 98.6\% for the ``rotten" class. In~\cite{method-fruitrec}, they adopted a bag-of-words model to locate fruits in images and combined several images from different views to estimate the number of fruits with a novel statistical model. This image processing method correlated 74.2\% between automatic counting numbers and the ground truth data. In~\cite{method-blueberry} and~\cite{method-blueberrydamage}, hyperspectral reflectance imaging methods were applied to determine the bruise or damage of blueberry. Pattern recognition algorithm was adopted to separate stem and calyx and detected blueberries with diseases and blueberries' orientations~\cite{method-blueberrydisease}. In~\cite{method-highspeedpotato}, they installed a 3-CCD line-scan camera and mirrors to capture a full view of potatoes. They also applied a method that combined Latent Dirichlet allocation (LDA) and a Mahalanobis distance classifier to detect the appearance defects of potatoes, and a Fourier-based shape classification method was utilized to detect misshapen potatoes as well.
In~\cite{method-gradingmango}, not only the proposed model realized grading the fruits with Multi-Attribute Decision Making (MADM), but also it successfully predicted the number of actual days that the harvested mangoes can be sent away with Support Vector Regression (SVR). In~\cite{deeprice}, they built a data set of rice -- FIST-Rice with 30,000 rice kernel samples and developed a system called Deep-Rice for grading rice by extracting the discriminative features from several angles of the rice. In~\cite{method-boiledshrimp}, they adopted Artificial Neural Networks (ANN) to classify the shapes of boiled shrimps by accepting the Relative Internal Distance (RID) values. The RIDs were calculated by segmenting the shrimp images and drawing the co-related lines on the segmented contour. The overall prediction accuracy of the ANN classifier is 99.80\%. In~\cite{2019tomato}, they presented a radial basis function SVM (RBF-SVM) model to detect the defects on Cherry and Heirloom tomatoes and developed a relation between the Lab color space of tomato images and the defects. The selected features for training and testing the model include color, texture, and shape features. In~\cite{PNCFapple}, they proposed a piecewise nonlinear curve fitting (PWCF) procedure to maintain and present the spectral features of the hyperspectral images, and the error-correcting output code-based support vector machine (ECOC-SVM) was adopted to address the apple bruise grading problem. In~\cite{rfid}, they utilized a Radio-frequency identification (RFID) tag as an electromagnetic transducer to convert the chemical-physical change of avocados into a modulation of the electromagnetic parameters when the avocados are ripening. The electromagnetic parameters were fed into a decision trees model to classify the ripeness level of avocados. Other than RFID, another radio frequency technology used is near-field communication (NFC) working with a colorimeter to overcome the inconsistent illumination conditions and improve the robustness of images from different devices. A colorimeter is a photosensitive instrument that can measure how much color an object or substance absorbs. It determines the color based on the red, blue, and green components of the light absorbed by the object or sample, similar to the human eye. In~\cite{nfc}, they integrated an NFC tag, a mobile phone NFC reader, and a colorimeter to detect the HSV (Hue, Saturation, Value) color space for fruits classification. In this research, hue and saturation were considered as the parameters to train the machine learning models. In~\cite{colorimetertomato}, they used the colorimeter to measure the chlorophyll fluorescence in tomatoes to determine the degree of ripeness. In~\cite{colorimeteravocado,colorimetermango}, they adopted colorimeters to extract the color information of avocado and mango, respectively, to access the ripeness and quality of each kind of fruit. The reviewed articles present that a colorimeter can interpret the red, blue, and green color of an object to avoid redundant color information and conquer inconsistent illumination conditions. However, such an instrument cannot represent an image's spatial information well due to missing color information. When spatial information is required to locate small objects, a colorimeter obviously cannot meet such needs, and a regular camera is still essential.

For banana grading, in~\cite{bananaann}, they developed a method to classify bananas into healthy and unhealthy groups based on image processing techniques and a neural network, and they obtained an accuracy of 97\%. In~\cite{banana2}, they designed a method to detect at which ripening stages red bananas are by measuring the dielectric properties of red bananas and sending the features to a Fuzzy~C-Means (FCM) classifier. In~\cite{bananafuzzy}, they also adopted a fuzzy model that was optimized with particle swarm optimization (PSO) technique to grade unripen, ripen and over-ripen bananas with the features of the bananas' peak hue and normalized brown area. The accuracy of this model is 93.11\%. A random forest classifier was utilized to grade the bananas according to the color features in~\cite{bananaml}, and the accuracy arrived at 94.2\%. In~\cite{bananaml2}, they also adopted machine learning algorithms to classify different types of bananas and their ripeness levels. SVM achieved an accuracy of 99.1\% to classify the banana types and has a 96.6\% accuracy in distinguishing the level of ripeness. In~\cite{bananann}, an ANN outperforms other machine learning algorithms to detect the ripeness of bananas with a feature vector that consists of color and texture features, and the classification accuracy of this system is 97.75\%. When it comes further adopting the IoT methods, \cite{edgefruit} proposed a framework to classify different date fruits by utilizing 5G and cloud.  The possibility of using cloud computing to detect apple chilling injury via spectrum-based classification is analyzed in~\cite{cloudfruit}.

Among the recently published articles, there is rarely research that can combine grading the fruits and locating the defective spots together. Compared to the reviewed related work, this study proposed a two-layer mechanism to realize both of the banana grading task and defective area locating mission and integrated edge computing and cloud computing into the system. A data set composed of 150 banana images was created, and these bananas were at different ripeness levels. Traditional data augmentation methods and a deep learning-based architecture called Cycle-Generative Adversarial Network (CycleGAN) were adopted to enlarge the data set to avoid overfitting. A feature vector containing color and texture features was used to train the first layer classifier. Next, the YOLOv3 model can detect the fruit peel's defected areas in the images in the ripened class, which is one of the first layer's output classes. This design illustrates the support of the Internet of Things structure for food processing and the entire controlling process of the endpoints (bananas on moving conveyor belts) via analyzing image classification results.

\section{Machine Vision System} \label{sec3}
\begin{figure*}[t!]
    \centering
    \includegraphics[scale=0.5]{ 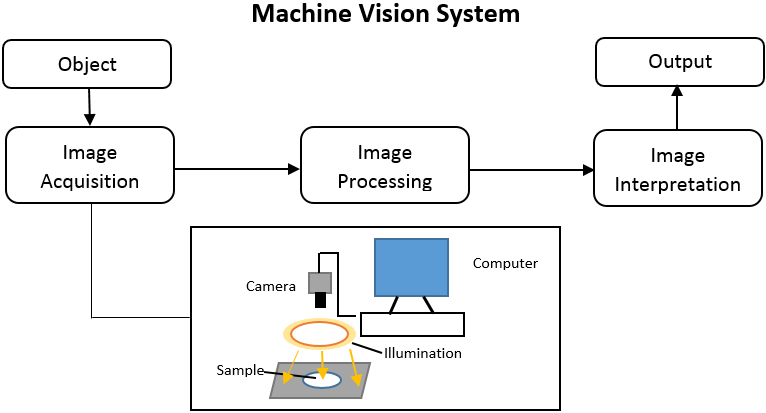}
    \caption{Principle components of a machine vision system.}
    \label{fig:mvs}
\end{figure*}

An MVS can observe, capture, assess, and recognize still or moving objects with one or several cameras automatically, usually in an industrial or production environment~\cite{mvs}. Then, the system utilizes the obtained data to control the following manufacturing procedures.

An MVS usually includes digital cameras, image processing programs, and a mechanical system, as shown in Fig.~\ref{fig:mvs}. The illumination device provides sufficient light to the object so that the camera can capture good quality images of the object. Then, the programs in the computer can process the images according to different purposes. The results of the processed images are for the mechanical system to make the next operation decision.

This work will focus on the image acquisition, image processing and image interpretation parts of the MVS.

\subsection{Image acquisition} \label{acquisition}
An MVS can obtain images in real-time via photographs, videos, and three dimensions (3D) techniques. There are many ways to acquire good quality images in food processing, such as stereo systems, remote sensing (RS), X-ray, thermal imaging, and Magnetic Resonance Imaging (MRI) \cite{stereosystem,rsforagri,satelliteremotesensing,xrayinagri,thermalimaging,mri}.

\subsection{Image processing}
\begin{figure*}[t!]
    \centering
    \includegraphics[width=\linewidth]{ 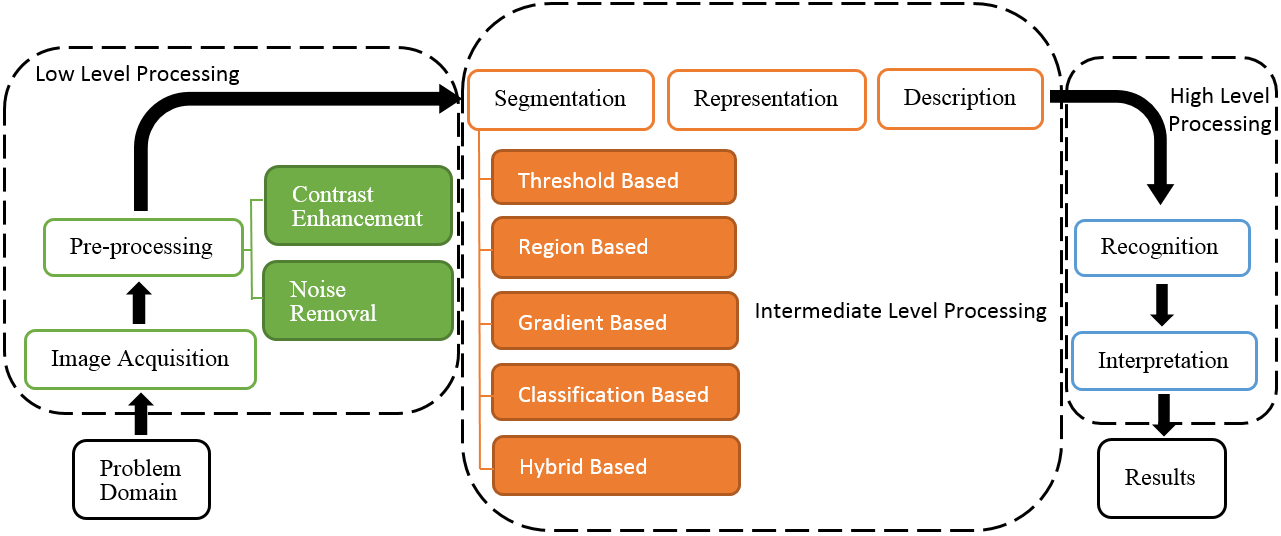}
    \caption{Different levels in image processing process.}
    \label{fig:imagelevels}
\end{figure*}

Image processing produces new images based on existing images to extract or improve the region of interest. This process is digital signal processing, and it does not involve interpreting the content or the meaning of the images. The different levels of the image processing process~\cite{cvforfood}, are shown in Fig.~\ref{fig:imagelevels}.

\subsubsection{Low level processing}
Low level processing contains image acquisition and image pre-processing. Image acquisition is the process of adopting different imaging equipment and sensing devices to obtain the images of samples and transfer them to a digital form that can be read by computers. Due to various imperfections in the shooting environment, such as insufficient illumination, long-distance or low resolution of imaging equipment, unstable viewfinder, and other impurities in the view, the original images usually require pre-processing for better performing image analysis. Typical image pre-processing methods include image enhancement such as adjusting the brightness or color of the images, cropping the images to focus on the region of interest, and noise removals such as undesirable noises or digital artifacts from low light levels. 

\subsubsection{Intermediate level processing}
Intermediate level processing includes image segmentation, image representation, and image description. Image segmentation is one of the essential steps in image processing, as it largely determines whether image analysis is focused on the target sample. Image segmentation is to separate the target from other useless image information so that the computational cost of subsequent image analysis can be reduced and improved accuracy. Boundary representation and region representation are both image representation. The previous one describes the size and shape features, while the latter is for the image's texture and defects. Image description can extract the quantitative information from the images which have already been processed by the previous steps.

\subsubsection{High level processing}
High level processing contains image recognition and image interpretation. During this step, statistical methods or deep learning methods are usually adopted to classify the target. These processes typically can determine how the following machines operate by serving useful information.

\subsection{Image interpretation}
The final step is image interpretation, where targets should be classified and useful spatial information from images should be derived. As a result, a decision based on the analysis result of the image can be made. Algorithms such as K-Nearest Neighbors, Support Vector Machine, neural networks, fuzzy logic, and genetic algorithms can help interpret the information obtained from the image. Neural network and fuzzy logic methods are proven to be involved with MVS in the food industry successfully~\cite{ann}.


\begin{figure*}[t!]
\centering

\subfigure[$G_{A2B}$ and $G_{B2A}$ are the two mappings between domains $A$ and $B$. $D_A$ and $D_B$ are the adversarial discriminators.]{
\begin{minipage}[t]{0.25\linewidth}
\centering
\includegraphics[scale=0.55]{ 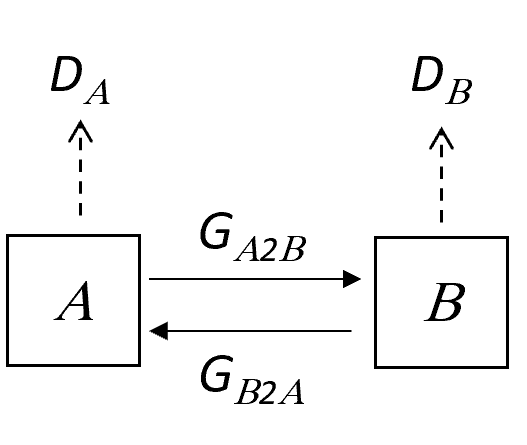}
\end{minipage}%
}%
\subfigure[forward cycle-consistency loss]{
\begin{minipage}[t]{0.35\linewidth}
\centering
\includegraphics[scale=1]{ 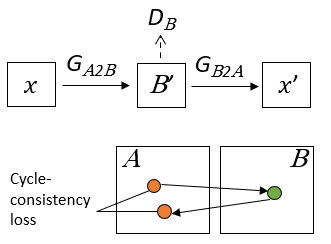}
\end{minipage}%
}%
\subfigure[backward cycle-consistency loss]{
\begin{minipage}[t]{0.35\linewidth}
\centering
\includegraphics[scale=1]{ 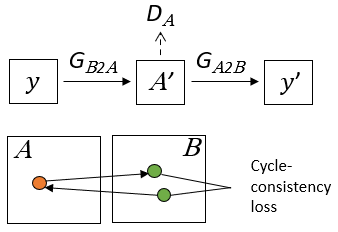}
\end{minipage}%
}%
                 
\centering
\caption{The principles of CycleGAN.}
\label{fig:cyclegan}
\end{figure*}

\section{Data set and Methodologies} \label{sec4}
This section describes the data set that was used followed by the methodologies.

\subsection{Data set}
In this study, the authors created a data set as the existing online open-access banana data sets only contain bananas in perfect conditions. The authors took 150 images of bananas at different ripeness levels and labelled them into three main groups: unripened, ripened, and over-ripened (50 images for each group). The ripened class has two sub-classes that are mid-ripened and well-ripened. The unripened group's bananas are still in green peels, while bananas in the ripened and over-ripened groups have a yellowish peel and different brown spots. However, 150 samples are not satisfactory for machine learning methodologies as it is easy to cause overfitting. As a result, the authors adopted data augmentation techniques, including traditional methods and a deep learning method -- CycleGAN, to enlarge the data set.

Traditional data augmentation methods such as rotation, flipping, and shifting are widely used for machine learning training. The authors also adopted CycleGAN to generate images of defective bananas. Generative adversarial net (GAN)~\cite{gan} is a generative model to learn the data distribution via an adversarial mode between a generating network and a discriminating network. The generating network generates samples similar to the real samples as much as possible, while the discriminating network tries to determine whether the samples are real samples or generated false samples. As illustrated in Fig.~\ref{fig:cyclegan}, CycleGAN~\cite{cyclegan} makes the principle of GAN apply to the image generation with ease. Based on GAN, CycleGAN adds another pair of the generator - discriminator, and cycle consistency loss to determine whether the generated images' style is consistent with the original data set images.

When there are two domains $A$ (style A) and $B$ (style B), $\{x_i\}^N_{i=1}$ and $\{y_j\}^M_{j=1}$ are samples where $x_i \in A$ and $y_j \in B$. The model involves a two mappings that are $G_{A2B}$ and $G_{B2A}$. $G_{A2B}$ is to transform the A style image to the B style image and vice versa for $G_{B2A}$. Additionally, two adversarial discriminators $D_{A}$ and $D_{B}$ are used to discriminate between the generated images and the real images. Therefore, if two GANs are being trained at the same time, one of the generator - discriminator pairs is $G_{A2B}$ and $D_{B}$ and the other pair is $G_{B2A}$ and $D_{A}$. Then, an image $x$ of style A should be able to transform back to itself after two transformations and image $y$ of style B is the same as described in Eq.~\ref{eq:gan}:

\begin{equation}
\begin{aligned}
G_{B2A}(G_{A2B}(x)) \simeq x, \\
G_{A2B}(G_{B2A}(y)) \simeq y.
\end{aligned}
\label{eq:gan}
\end{equation}

The first-order distance between the two graphs can be expressed as:

\begin{equation}
\begin{aligned}
L_{cyc}(G_{A2B}, G_{B2A}, A, B)&=\mathbb{E}_{x \sim A}[\parallel G_{B2A}(G_{A2B}(x)) - x \parallel _{1}]\\
&+\mathbb{E}_{y \sim B}[\parallel G_{A2B}(G_{B2A}(y))-y \parallel _{1}].
\end{aligned}
\label{eq:clc}
\end{equation}

Eq.~\ref{eq:clc} is the cycle consistency loss and Eq.~\ref{eq:term} is one of the terms of the total loss function:

\begin{equation}
\begin{aligned}
L(G_{A2B}, G_{B2A}, D_{A}, D_{B})&=L_{G}(G_{A2B}, D_{B}, A, B)\\
&+L_{G}(G_{B2A}, D_{A}, B, A) \\
&+\lambda L_{cyc}(G_{A2B}, G_{B2A}, A, B),
\end{aligned}
\label{eq:term}
\end{equation}

where $L_{G}(G_{A2B}, D_{B}, A, B)$ is the loss of $G_{A2B}$ and $D_{B}$ and $L_{G}(G_{B2A}, D_{A}, B, A)$ is the loss of $G_{B2A}$ and $D_{A}$. The expectation of CycleGan model is as Eq.~\ref{eq:exp}:

\begin{equation}
\begin{aligned}
G_{A2B}^{*}, G_{B2A}^{*} = & arg \min_{G_{A2B},G_{B2A}} \max_{D_{A},D_{B}}\\
&L(G_{A2B}, G_{B2A}, D_{A},D_{B}). \\
\end{aligned}
\label{eq:exp}
\end{equation}

\begin{figure}[t]
    \centering
    \includegraphics[scale=0.8]{ 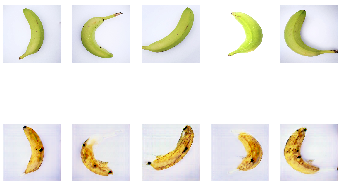}
    \caption{Five ripened banana images generated by CycleGAN (the second row) based on the unripened banana images (the first row). }
    \label{fig:gan}
\end{figure}

The comparison between the original unripened banana images and the generated ripened images is shown in Fig.~\ref{fig:gan}. The CycleGAN model created one hundred new ripened banana images. The total data set after data augmentation is as Table~\ref{tab:image}. 

\begin{table}[t]
\footnotesize
    \caption{The total data set after data augmentation.}
    \label{tab:image}
    \centering
    \begin{tabular}{c|c|c|c|c|c}
        \toprule
        Original&Rotation&Flipping&Shifting&CycleGAN&Total\\
        \midrule
        150&250&250&250&100&1000\\
        \bottomrule
    \end{tabular}
\end{table}

\subsection{Methodologies}
\begin{figure}[t]
    \centering
    \includegraphics[width=\linewidth]{ 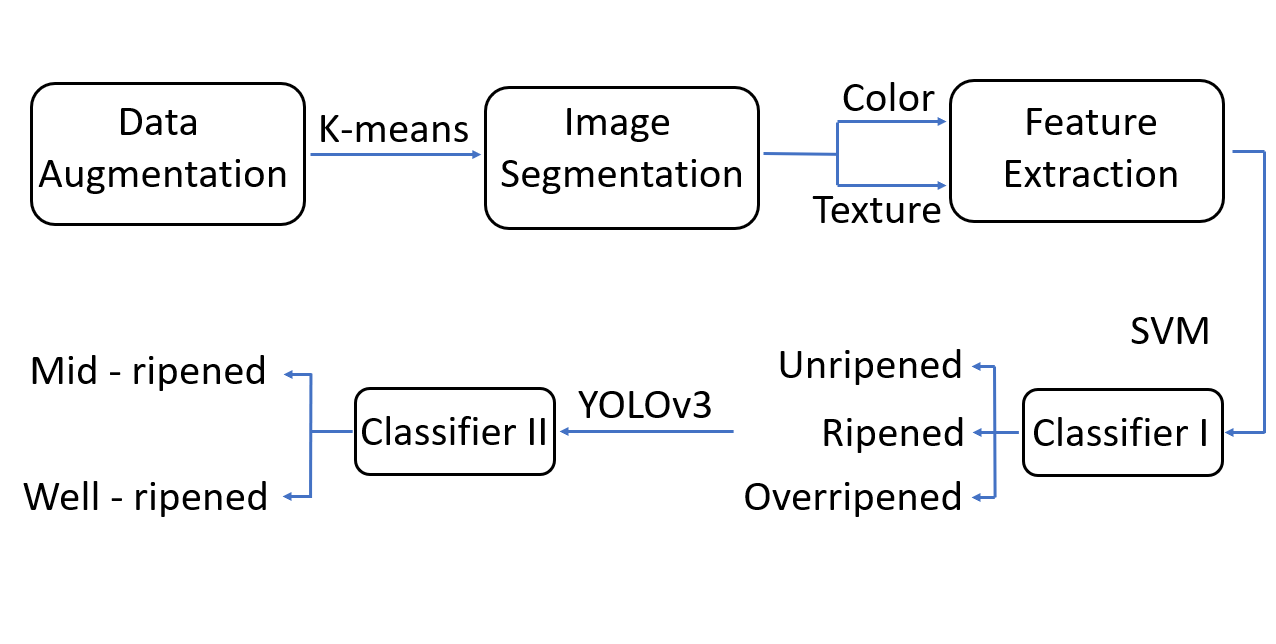}
    \caption{The flowchart of the proposed banana grading system.}
    \label{fig:flowchart}
\end{figure}

The proposed banana grading system includes data augmentation, image segmentation, feature extraction, and classification. The flowchart of the system is shown in Fig.~\ref{fig:flowchart}.

\subsubsection{Image segmentation}
\begin{figure}[t!]
    \centering
    \includegraphics[scale=0.4]{ 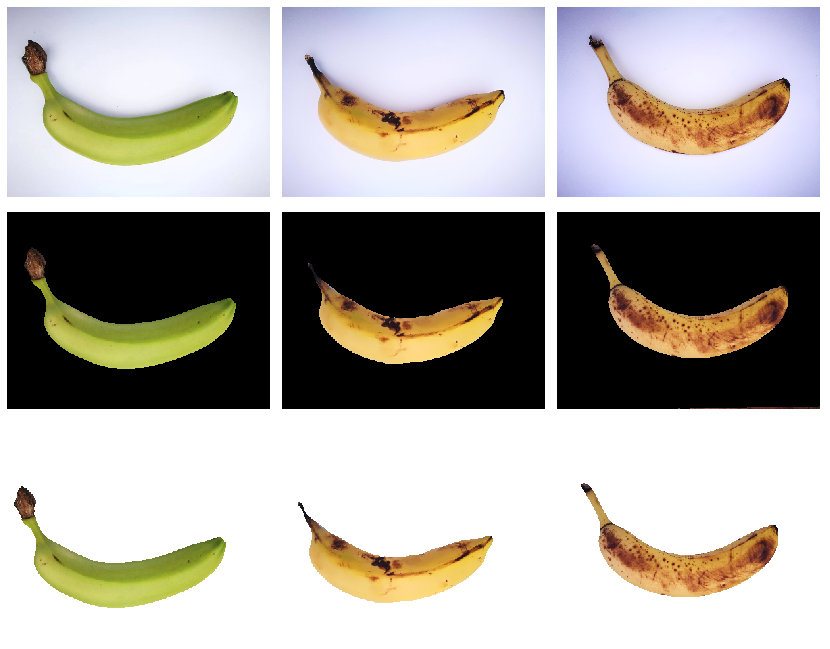}
    \caption{The sample inputs (the upper row), the masks (the middle row) and the outputs (the lower row) of image segmentation step.}
    \label{fig:seg}
\end{figure}

Image segmentation is to separate the target from the background in an image. Image segmentation is the first step of image analysis, the basis of computer vision, an important part of image understanding, and one of the most difficult problems in image processing. For a grayscale image, the pixels inside the region generally have intensity similarity but have discontinuous intensities at the region's boundaries. Methods for processing image segmentation include thresholding, region-based segmentation, edge detection-based algorithms, and machine learning-based methods. In this study, image acquisition was performed in natural light to achieve more real effects as in practical applications, resulting in inconsistent brightness in the background and shadows. Consequently, it is not easy to find a suitable threshold and complete and accurate edges to segment the target. Therefore, K-means is used here to address this problem. K-means utilizes distance as the evaluation index of similarity. The basic idea is to cluster the samples into different clusters according to the distance. The closer the two points are, the greater the similarity is. At the end of the process, all the data are allocated to the closest cluster center so that the sum of the squares of the distances between each point and its corresponding cluster center is minimized. Before applying K-means, rank filter and log transformation were adopted to reduce noise and improve image contrast. The sample segmentation results are shown in Fig.~\ref{fig:seg}.

\subsubsection{Feature Extraction}
For images, each image has its characteristics that can be distinguished from other types of images. Some are natural features that can be intuitively felt, such as brightness, edges, texture, and color. Some require transformation or processing to obtain, such as moments, histograms, and principal components. These features will be extracted in the form of numerical values or vectors so that the computer can identify images. Typical image features are color features, texture features, shape features, and spatial features.

\paragraph {Color Features}
Color feature is a global feature that describes the targets' surface properties corresponding to the image or image area. The general color feature is based on the pixels' characteristics, and all pixels that belong to the image or image area have their contributions. Color features can be extracted using methods such as color histograms, color sets, color moments, and color coherence vectors. In this study, since the unripened, ripened, and over-ripened bananas have distinctive color features (green, yellow, and brown), and unnecessary to consider the color space distribution, the color feature is one of the components that are extracted to train the classifier. Traditionally, RGB (Red, Greem, Blue) color space is prevalent in digital image processing. However, HSV color space is closer to how humans perceive color and more suitable for statistical analysis than RGB color space. Therefore, the color features of the proposed data set were extracted in the HSV color space. Eq.~\ref{eq:v}, Eq.~\ref{eq:s}, and Eq.~\ref{eq:h} can explain that how RGB color space converts to HSV color space. 

\begin{equation}
V = max (\frac{R}{255}, \frac{G}{255}, \frac{B}{255}),
\label{eq:v}
\end{equation}

\begin{equation}
S = 1 - \frac{3}{(R + G + B)} [min(R, G, B)],
\label{eq:s}
\end{equation}

\begin{equation}
H = \left\{
\begin{aligned}
\theta, ~~~G \geq B\\
2\pi - \theta,G < B,
\end{aligned}
\right.
\label{eq:h}
\end{equation}
~~~~~~~~~~~~~~~~~~~~~~~$where~\theta = cos^{-1}\bigg [ \frac{(R - G) + (R - B)}{2\sqrt{(R - G)^2 + (R - B)(G - B)}} \bigg ]$.

Due to the color characteristics of the three groups of bananas, the corresponding \textit{H}, \textit{S}, and \textit{V} value ranges are acquired from the analogy between HTML color codes and the natural colors of different banana peels~\cite{bananaann}. Table~\ref{tab:hsv} illustrates that \textit{H} and \textit{V} value ranges are distinct to be two of the input features to classify the bananas.

\begin{table}[t]
    \caption{The range for \textit{H, S, V} values in HSV color space for two banana groups.}
    \label{tab:hsv}
    \centering
    \begin{tabular}{|l|l|l|}
        \hline
        &Unripened&Ripened\\
        \hline
        \textit{H}&$72^\circ \leq H \leq 78^\circ$&$39^\circ \leq H \leq 72^\circ$\\
        \hline
        \textit{S}&$85\% \leq S \leq 100\%$&$70\% \leq S \leq 100\%$\\
        \hline
        \textit{V}&$27\% \leq V \leq 50\%$&$69\% \leq V \leq 100\%$\\
        \hline
    \end{tabular}
\end{table}

\paragraph{Texture Features}
The texture is another natural characteristic of the surface of an object. It describes the gray space distribution between the image pixels and the image area, and it will not change with different illumination. The texture feature is global, as well.  However, due to the texture is only a characteristic of an object's surface and cannot fully reflect the essential attributes of the object, it is impossible to obtain high-level image representation by only using texture features. Unlike color features, texture features are not pixel-based features. They require statistical calculations in an area containing multiple pixels. 

As a statistical feature, texture features often have rotation invariance and are robust to noise. Local Binary Pattern (LBP) ~\cite{lbp} is an operator used to describe the local texture features of an image, and it has significant advantages such as rotation invariance and gray invariance. The basic LBP operator is defined as a $3\times3$ size texture unit, and the value of the center pixel is the threshold. The grayscale value of the adjacent 8 pixels is compared with the center of the unit's pixel value. If the adjacent pixel value $g_0$ is greater than the center pixel value $g_c$, the pixel position is marked as 1; otherwise, it is 0. In this way, 8 pixels in the $3 \times 3$ unit can generate 8-bit binary numbers after compared to the center pixel. These 8-bit binary numbers are arranged in sequence to form a binary number. This binary number is the LBP value of the center pixel. Therefore, there are 256 LBP values, and the LBP value of the center pixel reflects the texture information of the area around the pixel. Mathematically, the process can be expressed as Eq.~\ref{eq:lbpthresh} and Eq.~\ref{eq:lbpbinary}, where $g_{c}$ is the center pixel value and $g_{0}$ is the adjacent pixel value. 

\begin{equation}
\label{eq:lbpthresh}
    s(g_{0} - g_{c}) = \left\{
    \begin{aligned}
    &1, g_{0} - g_{c} \geq 0 \\
    &0, g_{0} - g_{c} < 0,
    \end{aligned}
\right.
\end{equation}

\begin{equation}
\label{eq:lbpbinary}
    LBP = \sum_{p=0}^{7} s(g_{0} - g_{c}) 2^{p}.
\end{equation}

\begin{figure}[t]
    \centering
    \includegraphics[scale=0.5]{ 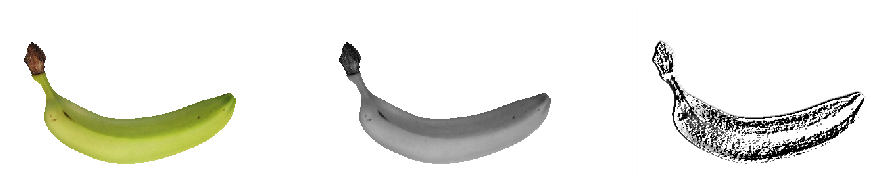}\\
    \includegraphics[scale=0.5]{ 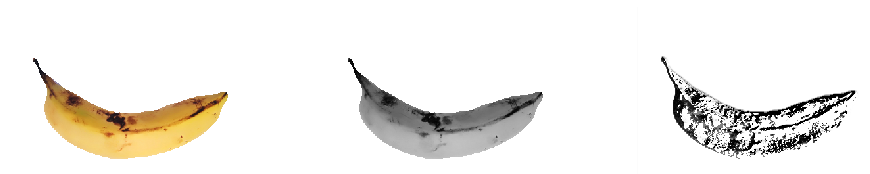}\\
    \includegraphics[scale=0.5]{ 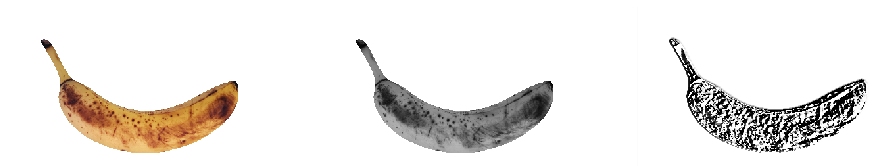}
    \caption{A sample image: (left) segmented image; (middle) grayscale image; (right) the texture feature that was extracted by the LBP operator.}
    \label{fig:lbp}
\end{figure}

Bananas at different ripeness levels can show distinctive texture features extracted by the LBP operator, as shown in Fig.~\ref{fig:lbp}.

\subparagraph{Shape Features.}
Various shape feature-based retrieval methods can more effectively use the region of interest in the image for retrieval. However, many shape features only describe the target's local nature, and a comprehensive description of the target often requires high computation time and storage capacity. Requirements. Moreover, the target shape information reflected by many shape features is not entirely consistent with human perception because the shape reflected from the 2-D image is often not the actual shape of the 3-D object due to the change of viewpoint and distortion.

\subparagraph{Spatial Features.}
Spatial features refer to the mutual spatial position or relative orientation relationships between multiple target segments in the image. These relationships can also be divided into adjacent, overlapping, and contained. Generally, the spatial position information can be divided into relative spatial position information and absolute spatial position information. The former relationship emphasizes the relative situation between the targets, including horizontal and vertical positions. The latter relationship emphasizes the distance and orientation of the targets. The relative spatial position can be deduced from the absolute spatial position, but it is often relatively simple to express the relative spatial position information. Spatial relationship features can enhance the ability to describe and distinguish image content, but spatial relationship features are often more sensitive to the image, object rotation, inversion, and scale changes. Besides, in practical applications, only using spatial information is often insufficient to effectively and accurately express scene information. Usually, other features are needed to cooperate with spatial relationship features.

~\\
 \indent In this study, two feature vectors, which were $A = [H~V~LBP]$ and $B = [H~V]$, were input into the candidate models respectively to select which feature vector would yield the optimal result.

\subsubsection{Classification}
In this study, the classification task is divided into two steps. The first step is to feed extracted features into a traditional machine learning classifier to separate bananas from unripened, ripened, and over-ripened groups as traditional machine learning methods usually have simple architecture. They will not require too many computational resources. The authors applied four models in this research, which are K - Nearest Neighbours (KNN), Random Forest (RF), Naive Bayes (NB), and Support Vector Machine (SVM), and compared the performances of these four models. These four models are efficient and straightforward. They can handle high-dimensional data and do not need to make feature selection. SVM, RF, and BN have low computational overhead, and KNN performs better than SVM on simple classification problems. Therefore, these four models are considered candidates in this study. The brown spots on banana peels will not be detected here as there is no consistency of the level of brown color that should be included, and for over-ripened bananas, the detection of multiple irregular areas caused by the connected brown areas will result in inaccurate results.
Additionally, it is unnecessary to detect the brown spots for over-ripened bananas because the peel is mainly brown. As a result, the bananas will be classified into three primary groups. The next step is to feed the output ripened fruit images from SVM into the YOLOv3~\cite{redmon2018yolov3} transfer learning model to detect the brown spots and separate the bananas into mid-ripened and well-ripened groups according to how many brown areas they have.

\begin{figure}[t]
    \centering
    \includegraphics[width=\linewidth]{ 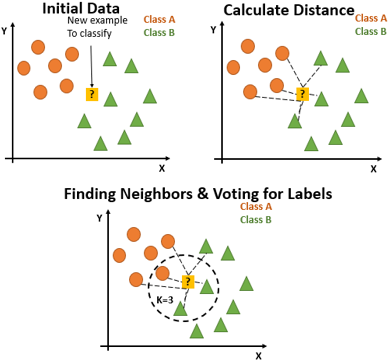}
    \caption{An example of KNN algorithm.}
    \label{fig:KNN}
\end{figure}

\paragraph{K - Nearest Neighbours}
K-nearest neighbour method is one of the basic machine learning methods. Its implementation method is to input test data into the model trained by the data and labels in the training set. The test data features are compared with the corresponding features in the training set, and the first \textit{K} samples in the training set that are most similar to it are found. The category corresponding to the test data is the category that appears most frequently among the \textit{K} samples.  KNN algorithm is illustrated in Fig.~\ref{fig:KNN}. This principle also shows that the result of the KNN algorithm mainly depends on the choice of the \textit{K} value. 

In KNN, the distance between samples is calculated as an index of dissimilarity between samples, avoiding matching between samples. Generally, Euclidean distance (Eq.~\ref{Euclidean}) and Manhattan distance (Eq.~\ref{Manhattan}) are the most popular methods to calculate the distance in KNN.
\begin{equation} \label{Euclidean}
\begin{aligned}
    D(x,y)&=\sqrt{(x_1-y_1)^2+(x_2-y_2)^2+...+(x_n-y_n)^2}\\
    &=\sqrt{\sum_{i=1}{n}(x_i-y_i)^2},
\end{aligned}
\end{equation}

\begin{equation} \label{Manhattan}
\begin{aligned}
    D(x,y) &= |x_1-y_1|+|x_2-y_2|+...+|x_n-y_n|\\
    &=\sum_{i=1}{n}|x_i-y_i|.
\end{aligned}
\end{equation}

At the same time, KNN makes decisions based on the dominant category of \textit{K} samples, rather than a single sample category decision. These two points are the advantages of the KNN algorithm.

\begin{figure}[t]
\centering
\subfigure[An example of decision tree.]{
\begin{minipage}[t]{\linewidth}
\centering
\includegraphics[width=\linewidth]{ 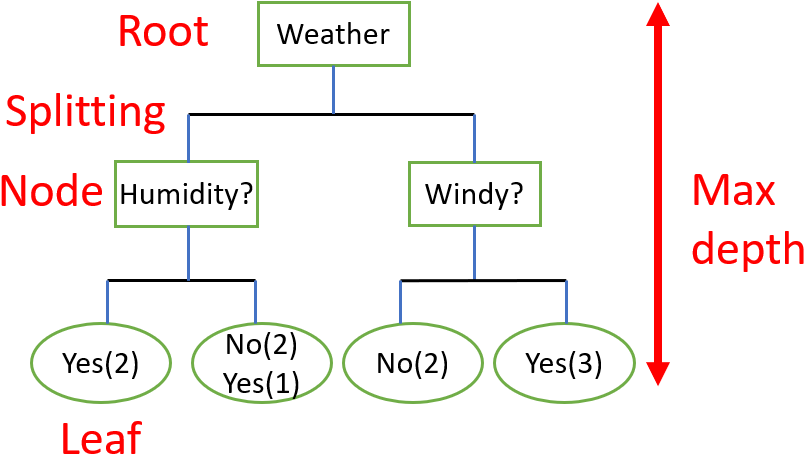}
\label{fig:dt}
\end{minipage}%
}%
\\
\subfigure[The explanation of random forest.]{
\begin{minipage}[t]{\linewidth}
\centering
\includegraphics[width=\linewidth]{ 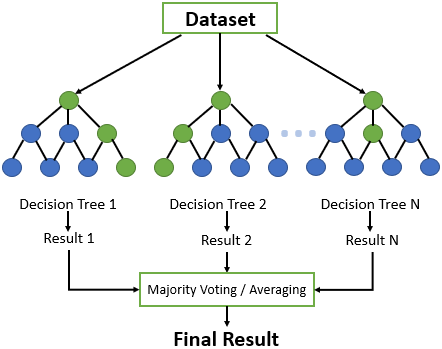}
\label{fig:rf}
\end{minipage}%
}%
\centering
\caption{Decision tree and random forest.}
\end{figure}

\paragraph{Random Forest}
Before explaining random forests, the concept of decision trees needs to be introduced first. A decision tree is a straightforward algorithm. The analyzing process and results are explainable and also in line with human intuitive thinking. The decision tree is a supervised learning algorithm based on if-then-else rules, and these rules are obtained through training rather than manual formulation. The logic of the decision tree is shown in Fig.~\ref{fig:dt}.

Random forest is composed of many decision trees, and there is no correlation between these decision trees. When performing a classification task, new input samples enter, and each decision tree in the forest determines its decision independently. Each decision tree will get its classification result. The most appeared decision in all the classification results will be considered as the final result. The relation between decision tree and random forest is shown in Fig.~\ref{fig:rf}.

\paragraph{Naive Bayes}
Bayesian classification is a general term for a class of classification algorithms. These algorithms are based on Bayes' theorem, so they are collectively called Bayesian classification. Naive Bayesian classification is the simplest and most common classification method in Bayesian classification. Its idea is to calculate the posterior probability of the \textit{Y} variable belonging to a specific category based on individual prior probabilities. Eq.~\ref{nb} illustrates that the probability of \textit{Y} occurring under \textit{X} conditions can be determined by knowing the three parts in the right-hand side of the equation, which are the probability of \textit{X} event ($P(X)$, the prior probability of \textit{X}), the probability of \textit{Y} belonging to a specific class ($P(Y)$, the prior probability of \textit{Y}), and the probability of event X under a particular category of known \textit{Y} ($P(X|Y)$, the posterior probability).

\begin{equation} \label{nb}
    P(Y|X)=\frac{P(YX)}{P(X)}=\frac{P(X|Y)P(Y)}{P(X)}.
\end{equation}

\begin{figure}[t]
    \centering
    \includegraphics[width=\linewidth]{ 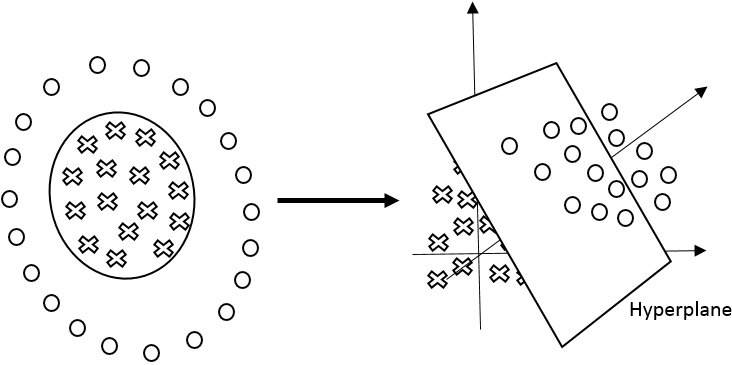}
    \caption{Nonlinear data classification in high-dimensional feature space with the SVM.}
    \label{fig:svm}
\end{figure}

\paragraph {Support Vector Machine}
Cortes and Vapnik proposed SVM in~\cite{svm} that is a supervised learning method and can be widely applied to statistical classification and regression analysis. Its basic model is a linear classifier defined on the feature space to find the hyperplane with the maximum interval between two types of data. The learning strategy of SVM is to maximize the interval, which can be formalized as a problem to solve the convex quadratic programming, which is also equivalent to the minimization problem of a regularized hinge loss function. However, the data is not linearly separable for most of the time. Under this circumstance, the hyperplane that meets the condition does not exist at all. For nonlinear situations, the SVM approach is to choose a kernel function. The SVM first completes the calculation in the low-dimensional space and then maps the input space to the high-dimensional feature space through the kernel function. Finally, the optimal separating hyperplane is constructed in the high-dimensional feature space so that the nonlinear data are separated, as shown in Fig.~\ref{fig:svm}. For multi-classes tasks, a nonlinear SVM with Radial Basis Function (RBF) kernel can be applied. The RBF kernel is $k(x^{(i)},x^{(j)})=exp(-\gamma\|x^{(i)}-x^{(j)}\|^{2})$. 

\paragraph {YOLOv3}
You Only Look Once (YOLO)~\cite{yolo} is an object recognition and localization algorithm based on deep neural networks. The most distinct feature of YOLO is that it runs fast and can be used in real-time systems. However, its mean Pixel Accuracy (mPA) towards small objects is not satisfactory. On the premise of maintaining YOLO's speed advantage, YOLOv3 adopted the residual network structure to form a deeper network level and uses multi-scale features for object detection. Also, object classification uses Logistic instead of softmax, which improves prediction accuracy, especially for small object recognition capabilities. In YOLOv3, there are only convolution layers, and the size of the output feature map is controlled by adjusting the convolution step. Therefore, there is no particular limitation on the size of the input picture. YOLOv3 draws on the idea of Feature Pyramid Networks (FPN) -- small size feature maps are used to detect large-sized objects while large-sized feature maps are used to detect small-sized objects. Comparing to the other structures that are prevalent for object detection, such as Single Shot MultiBox Detector (SSD), Faster Region-based Fully Convolutional Networks (Faster R - FCN), and RetinaNet, the inference time of YOLOv3 is significantly faster~\cite{redmon2018yolov3}. This advantage is of great importance in the industry as the high detection accuracy is valuable and the detection speed also plays an important role. In this research, the YOLOv3 model satisfies the need to detect and locate the small spots on banana peels and offers a fast response to the inline application. Here, when the YOLOv3 model detects five or less defected areas, this sample will be considered a mid-ripened sample. Also, a well-ripened sample is determined by whether there are more than five defected areas found by the model.

\subsubsection{Cloud Computing and Edge Computing}
\paragraph{Cloud Computing}
Cloud computing is a computing resource delivery model that integrates various servers, applications, data and other resources and provides these resources in the form of services through the Internet \cite{xu2012cloud}. Cloud computing services usually run on a global network of secure data centers, regularly upgraded to the latest fast and efficient computing hardware. Compared with a single organization data center, it can reduce applications' network latency and improve economic efficiency. Simultaneously, it can simplify data backup, disaster recovery, and business continuity at a lower cost as data can be mirrored on multiple redundant sites in the cloud provider's network \cite{ahmed2014cloud}. In the training process of deep learning, making learning effective requires a large quantity of data. The deep learning architecture ensures multiple levels of neural networks. When the depth (the number of layers) is greater, more storage space is needed for the large amount of data required for training. As tasks become computationally intensive, power requirements will increase. Therefore, traditional computers may not operate efficiently. This also leads to more capital investment in research and development institutions. Therefore, performing depth learning training and analysis in the cloud has become an ideal, simple and effective method~\cite{cloudlearning}.

\begin{figure}[t]
    \centering
    \includegraphics[width=\linewidth]{ 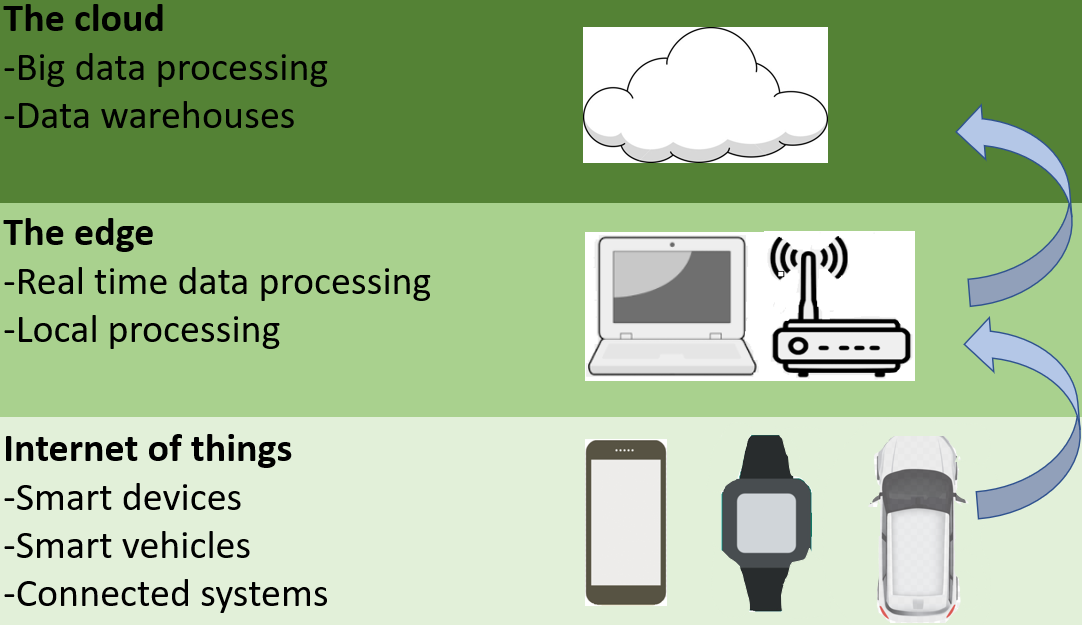}
    \caption{Edge computing architecture.}
    \label{fig:edgecloud}
\end{figure}

\paragraph{Edge Computing}
In the field of the Internet of Things, the edge refers explicitly to the end device's vicinity, so edge computing is the computation generated near the end device. The network's edge can be any functional entity from the data source to the cloud computing center. These entities are equipped with an edge computing platform that integrates the network's core capabilities, computing, storage, and applications, providing end-users with real-time, dynamic and intelligent service computing \cite{edgelearning}. Unlike processing and algorithmic decision-making of cloud computing, which needs to be performed in the cloud, edge computing is an action that pushes intelligence and computing closer to reality. The main differences between cloud computing and edge computing are reflected in multi-source heterogeneous data processing, bandwidth load and resource waste, resource limitation, and security and privacy protection. Therefore, the significant problems that edge computing is solving are the high latency, network instability, and low bandwidth problems that exist in the traditional cloud computing mode due to resource constraints. By migrating some or all of the processing procedures to be close to users or data collection points, edge computing can significantly reduce the impact on applications in cloud-centric sites \cite{pan2017future}. The general structure of edge computing is illustrated in Fig.~\ref{fig:edgecloud}.

\subsubsection{Method Evaluation}
The accuracy (Eq.~\ref{eq:acc}), sensitivity/recall (Eq.~\ref{eq:sensitivity}), precision (Eq.~\ref{eq:specificity}) and F1 - score (Eq.\ref{eq:f1}), all of which are common evaluation methods in statistics, are used to evaluate the first-layer classification results.

\begin{equation}
\begin{aligned}
   Accuracy = \frac{TP + TN}{TP + TN + FP + FN},\\
   \end{aligned}
    \label{eq:acc}
\end{equation}

\begin{equation}
\begin{aligned}
   Sensitivity/Recall = \frac{TP}{TP + FN},\\
   \end{aligned}
    \label{eq:sensitivity}
\end{equation}

\begin{equation}
\begin{aligned}
   Precision = \frac{TP}{TP + FP},\\
   \end{aligned}
    \label{eq:specificity}
\end{equation}

\begin{equation}
\begin{aligned}
   F1 - score = 2 \cdot \frac{Precision \cdot Reccall}{Precision + Reccall}.\\
   \end{aligned}
    \label{eq:f1}
\end{equation}
  
$\begin{aligned}
&TP = True~positive, TN = True~negative, \\
&FP = False~positive, FN = False~negative.
\end{aligned}$

\begin{figure*}[t!]
    \centering
    \includegraphics[width=\linewidth]{ 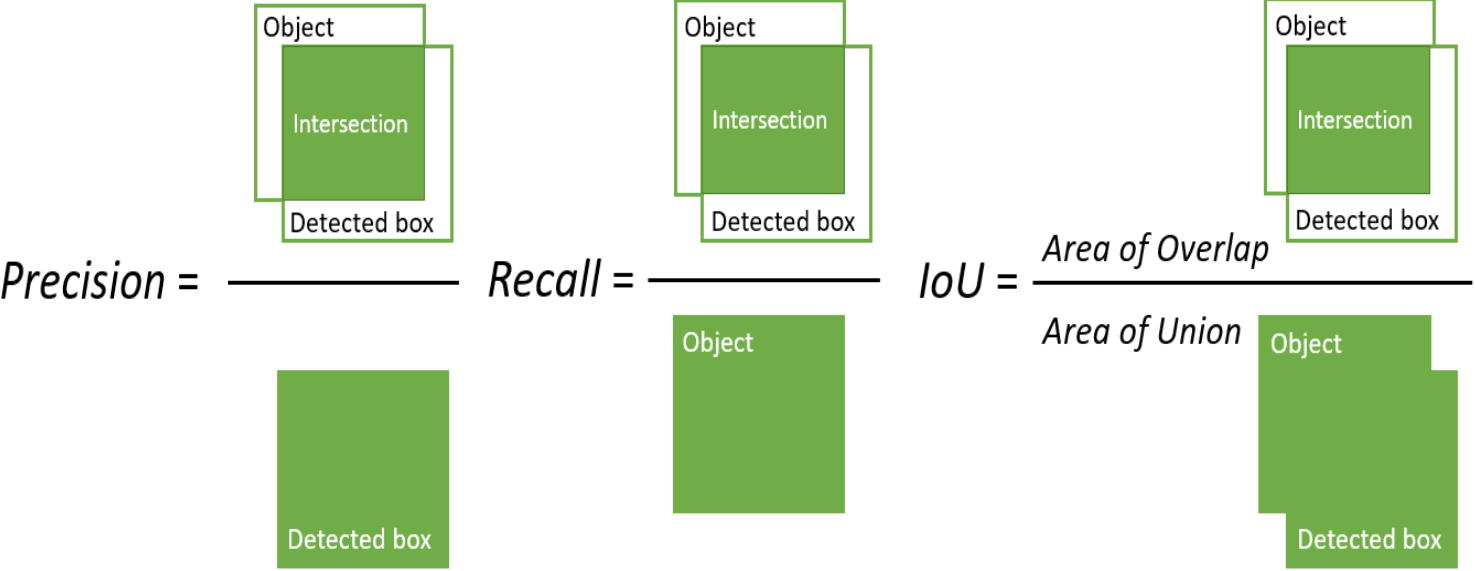}
    \caption{The definitions of the evaluation methods.}
    \label{fig:eva}
\end{figure*}

\vspace{1.5\baselineskip}
To further assess the performance of the YOLOv3 model, mAP (mean Average Precision), Intersection over Union (IoU), and recall are applied to evaluate the predicted locations of the target. The definitions of the evaluation methods are shown in Fig.~\ref{fig:eva}. 

The area of overlap means the area in both the ground truth bounding box and the predicted bounding box, while the area of union is the total area of the ground truth bounding box and the predicted bounding box as:

\begin{equation}
\label{eq:iou}
    IoU = \frac{Area~of~Overlap}{Area~of~Union}.
\end{equation}

\section{Experiments Results and Discussion} \label{sec5}
The experiments were conducted on the Ubuntu 18.04.4 LTS system, with an Intel$^\circledR$ Core$^{\text{TM}}$ i7-8700K CPU @ 3.70GHz × 12 processor, 32G memory, and GeForce GTX 1080 Ti/PCIe/SSE2 graphic. As a cloud server, we used a local available server at our laboratory.

For the first classification layer, 800 images were used to train the first layer classifier, and 200 images were for testing. After forming the the feature vector $A = [H~V~LBP]$ with the \textit{H} value, \textit{V} value, and LBP features, and the feature vector $B = [H~V]$ with only the \textit{H} value and \textit{V} value, the extracted $A$ and $B$ of all the training images were fed into the four classifiers for training, respectively. For the first layer classifier, the performances between KNN, RF, NB, and SVM were tested in this step. Table~\ref{tab:1st} presents the results of different methods training with feature vector $A$. The results show that the training times of RF and NB are significantly less than the training times of KNN and SVM, however, the performance of SVM training with feature vector $A$ outweighs the other algorithms. Additionally, when the KNN model training with vector $B$, the accuracy is noticeable lower than when the model training with vector $A$, while the other models show no significant difference between training with $A$ and $B$. 

\begin{table*}[t!]
\footnotesize
\newcommand{\tabincell}[2]{\begin{tabular}{@{}#1@{}}#2\end{tabular}}
    \caption{The comparison between different algorithms as the first layer classifier (all models were trained with feature vector $A$/$B$).}
    \label{tab:1st}
    \centering
    \begin{tabular}{|l|c|c|c|c|}
        \hline
        Method&Parameters&\tabincell{l}{Preprocessing Time\\(second/per image)}&\tabincell{l}{Training Time\\(second)}&Accuracy \\
        \hline
        KNN + $A$&\multirow{2}*{$K = 3$}&\multirow{6}*{\tabincell{l}{0.093 with $A$\\0.081 with $B$}}&51.313&92.87\%\\
        \cline{1-1}\cline{4-5}
        KNN + $B$&~&~&50.717&89.42\%\\
        \cline{1-2} \cline{4-5}
       RF + $A$&\multirow{2}*{100 Trees}&~&5.769&96.80\%\\
        \cline{1-1} \cline{4-5}
        RF+ $B$&~&~&5.554&94.57\%\\
        \cline{1-2} \cline{4-5}
        NB + $A$&\multirow{2}*{--}&~&1.396&94.50\%\\
        \cline{1-1} \cline{4-5}
        NB + $B$&~&~&1.393&94.15\%\\
        \cline{1-2} \cline{4-5}
        SVM + $A$ &\multirow{2}*{\tabincell{l}{\textit{g}=0.005,\\ \textit{C}=1000}}&~&111.657&98.50\%\\
        \cline{1-1} \cline{4-5}
        SVM + $B$ &~&~&111.158&97.63\%\\
        \hline
    \end{tabular}
\end{table*}

As the SVM performs better than the other models, training the SVM classifier with the entire features was conducted to test if the extracted feature vector $A$ is the most effective one. Table~\ref{tab:1st-ext} shows the performance of training with the feature vector $A$ is superior to training with the entire features in both accuracy and processing time aspects. 

\begin{table}[t!]
\footnotesize
    \caption{The comparison between applying extracted features ($A$) and entire features.}
    \label{tab:1st-ext}
    \centering
    \begin{tabular}{|l|c|c|}
        \hline
       Method&SVM+$A$&SVM+entire features\\
        \hline
        Accuracy&98.50\%&95.78\%\\
        \hline
        Parameter&\multicolumn{2}{c|}{\textit{g}=0.005, \textit{C}=1000}\\
        \hline
    \end{tabular}
\end{table}

Table~\ref{tab:svm} shows the confusion matrix for the SVM's testing result. This confusion matrix demonstrates that the overall predicting accuracy of the SVM classifier achieved 98.50\% (with \textit{g} = 0.005 and \textit{C} = 1000). The three mispredictions happened in the ripened class and the over-ripened class. The reason for the wrong predictions is that some images in the ripened class and over-ripened class are quite similar and the labelling was finished manually. As a result, for those images that also can be labelled as another class, it is difficult for the classifier to distinguish.

\begin{table}[t!]
\scriptsize
    \caption{The confusion matrix for the first-layer classifier.}
    \label{tab:svm}
    \centering
    \begin{tabular}{|l|c|c|c|c|}
        \hline
        ~&\multicolumn{3}{c|}{\textbf{Predicted Class}}&\multirow{2}{*}{Sensitivity} \\
        \cline{1-4}
        \textbf{True Class}&Unripened&Ripened&Overripened&\\
        \hline
        Unripened&66&0&0&100\\
        \hline
        Ripened&0&60&2&96.77\%\\
        \hline
        Overripened&0&1&71&98.61\%\\
        \hline
        Precision&100&98.36\%&97.26\%&Acc=98.50\%\\
        \hline
    \end{tabular}
\end{table}

\begin{figure}[t!]
\centering

\subfigure[A mid-ripen sample.]{
\begin{minipage}[t]{0.45\linewidth}
\centering
\includegraphics[height=2cm]{ 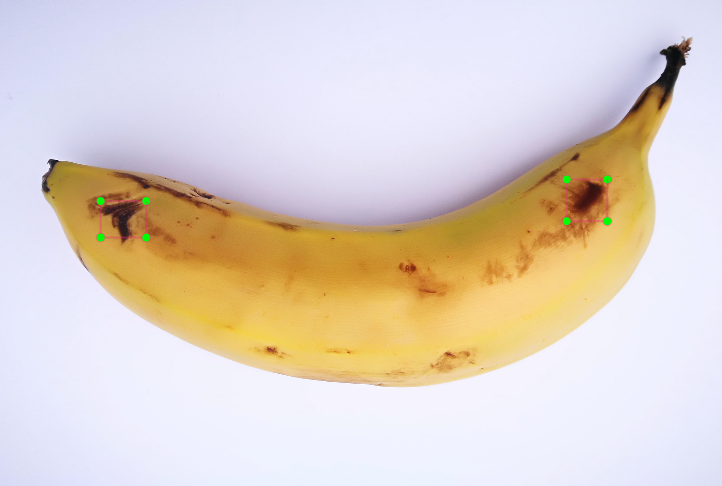}
\end{minipage}%
}%
\subfigure[A well-ripen sample.]{
\begin{minipage}[t]{0.45\linewidth}
\centering
\includegraphics[height=2cm]{ 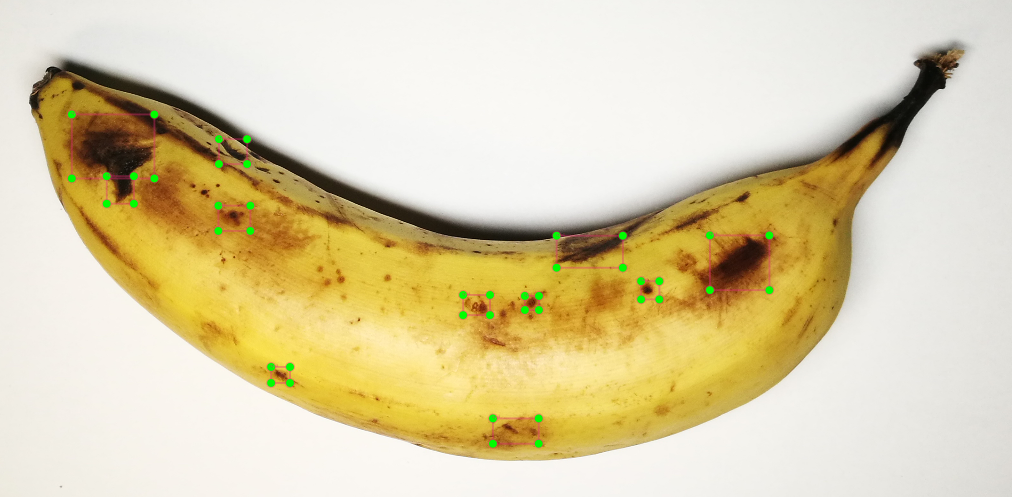}
\end{minipage}%
}%
                 
\subfigure[A predicted mid-ripen sample.]{
\begin{minipage}[t]{0.45\linewidth}
\centering
\includegraphics[height=2cm]{ 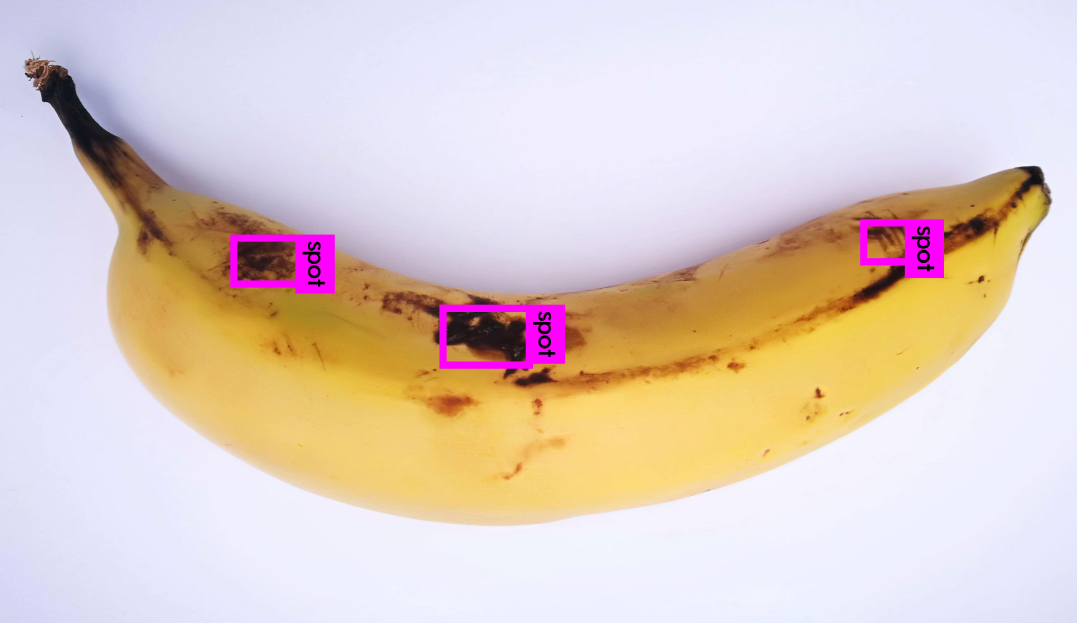}
\end{minipage}
}%
\subfigure[A predicted well-ripen sample.]{
\begin{minipage}[t]{0.45\linewidth}
\centering
\includegraphics[height=2cm]{ 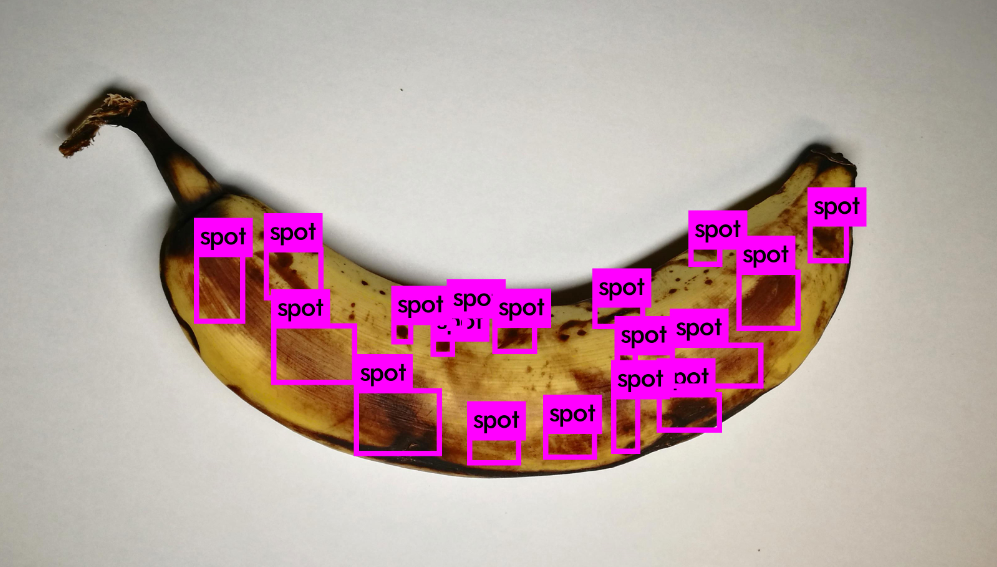}
\end{minipage}
}%

\centering
\caption{The sample ground truth data and the predicted samples in the mid-ripen and well-ripen classes.}
\label{fig:label}
\end{figure}

For the second layer, the ripened group's defective areas were labelled with an open-source tool called ``labelImg"~\cite{label} manually, and the 61 images that were predicted as ripened were fed to the second predictor. One sample ground truth data from each class is shown in Fig.~\ref{fig:label}. However, all the ground truth data were labelled based on subjective judgment, which will affect the criteria of being in which class. Therefore, an objective standard of banana ripeness levels should be referenced in the following work.

After 10,000 iterations and 40 hours of training, the mAP of the testing results is 0.8239, and the average IoU is 76.94\%, the average recall and precision of the testing results are 91.77\% and 75.28\%, respectively. The average processing time to predict a testing image is 0.053 seconds. The high recall and low precision could be that the model detected some spot areas that were not labelled on the ground truth data. According to the detected areas' results, to which sub-class this sample belongs will be determined by the number of the detected areas. When the detected areas are more than five, this banana will be classified as the well-ripened group. The IoU result indicates that the predicted areas shifted from the ground truth labels to some extent, but this will not affect the predicted areas. As a result, the confusion matrix, shown in Table~\ref{tab:2nd}, is based on the number of predicted areas is still valid.

\begin{table}[t!]
    \scriptsize
    \caption{The confusion matrix for the second-layer classifier.}
    \label{tab:2nd}
    \centering
    \begin{tabular}{|c|c|c|c|}
        \hline
        ~&\multicolumn{2}{c|}{\textbf{Predicted Class}}&\multirow{2}{*}{Sensitivity} \\
        \cline{1-3}
        \textbf{True Class}&Mid-ripened&Well-ripened&\\
        \hline
        Mid-ripened&31&2&93.94\%\\
        \hline
        Well-ripened&4&26&86.67\%\\
        \hline
        Precision&88.57\%&92.85\%&Acc=90.16\%\\
        \hline
    \end{tabular}
\end{table}

\section{Application} \label{sec6}
\begin{figure*}[t!]
    \centering
    \includegraphics[width=\linewidth]{ 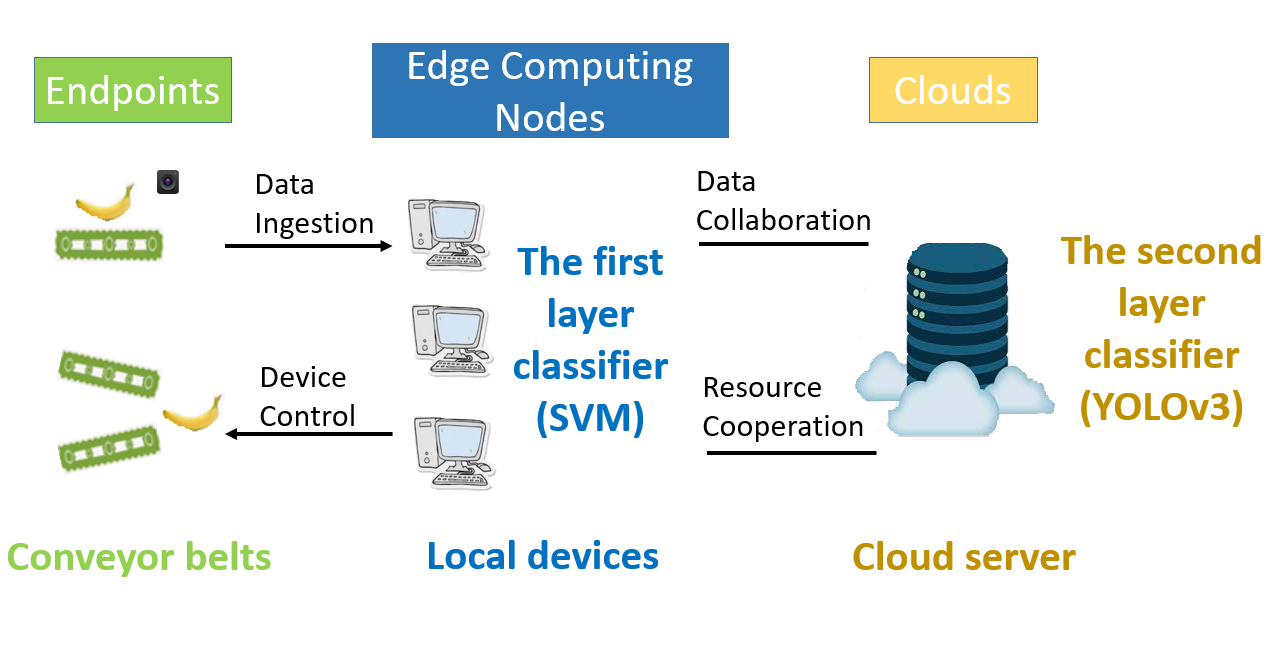}
    \caption{The structure of the banana grading system.}
    \label{fig:system}
\end{figure*}

In this section, a proposed system for banana grading will be presented in detail. This system contains three layers: mechanical layer, local layer, and cloud layer, shown in Fig.~\ref{fig:system}. The proposed mechanical layer will be set up in future work since this research is focused on the classifier models and the GUI.

The mechanical layer has two functions. At the beginning of the system, a conveyor belt can transfer food to a tunnel with a steady illumination source and a camera hanging on the top of it. In this tunnel, the camera can take pictures of the food. Then, the image will be sent to the local layer. The other function of the mechanical layer works at the end of the system as there is a switch that can guide the food to two tracks. One of the tracks is for healthy food, while the other one is for defective food. Once the image of the food is sent to the local layer, the local computer will execute pre-processing to the image, such as contrast enhancement and cropping, so that it will be possible to gain an accurate analyzing result. 

The first layer of the proposed two-layer grading model will be placed on the local layer, but the second layer will be implemented on the cloud layer. Since the first layer of the model is SVM, a traditional machine learning model, and it does not cost too much computational resource, placing it on edge (the local side) will improve the device's response time. When the image does not need to be analyzed by the second layer of the model, edge computing can reduce the data flow from the device to the cloud. Meanwhile, cloud computing power still plays a role when further image analysis is required. This arrangement will also be useful when the application is transplanted to a smartphone application in future research. If the SVM model's result needs to be further confirmed its grade, the image will be sent to the YOLOv3 model, which is implemented on the cloud server. This deep learning model needs to run on a GPU-based system; however, not every food processing warehouse has this condition. Therefore, a GPU-based cloud server is capable of tackling this problem. After the model on the cloud assesses the image, the result will be sent back to the local layer. According to the result, the system will control the switch on the conveyor belt to enable the food to move to the right track.

\begin{figure}[t!]
    \centering
    \includegraphics[width=\linewidth]{ 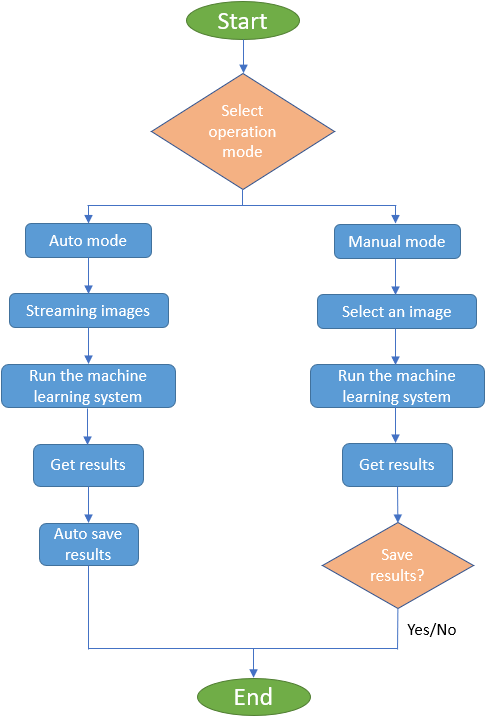}
    \caption{The flow chart of the application.}
    \label{fig:appflowchart}
\end{figure}

At last, a user-side application with a user-friendly GUI is also included in this system. The flow chart of how the application works is shown in Fig.~\ref{fig:appflowchart}. This application can be used in an auto mode or a manual mode. In auto mode, the inline camera hanging above the conveyor belt will stream the banana images to this application. Then the application will process each image and save the results automatically. With the manual mode, the user can assess images by taking pictures of bananas manually as sometimes it is not necessary to evaluate every single banana.

\begin{figure}[t!]
\centering
\subfigure[An unripened banana.]{
\begin{minipage}[t!]{0.85\linewidth}
\centering
\includegraphics[width=\linewidth]{ 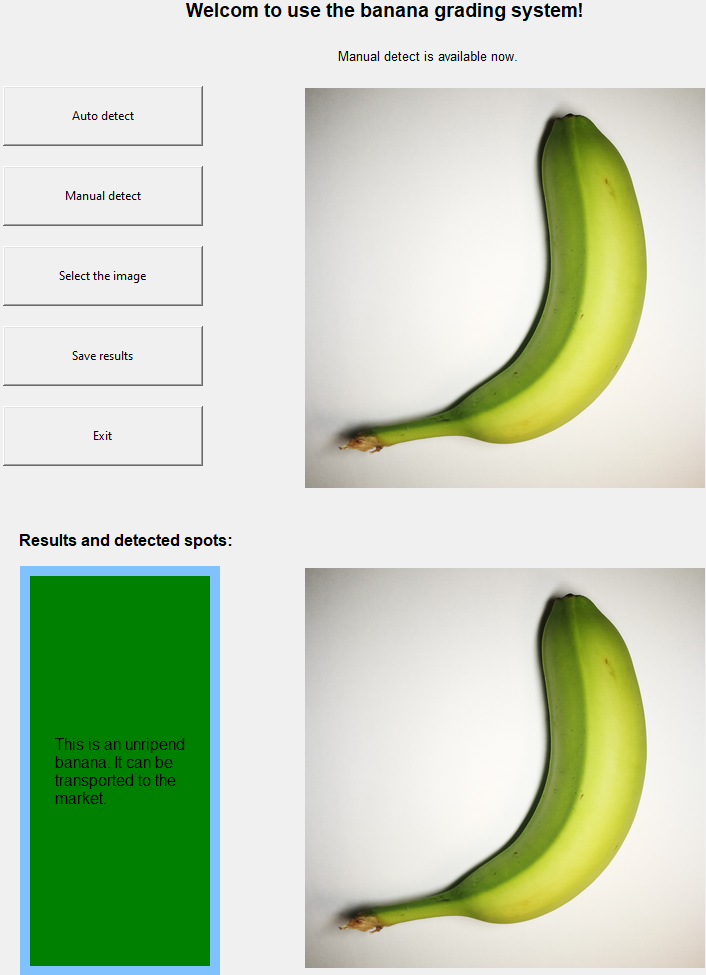}
\label{fig:app-unrippened}
\end{minipage}%
}%
\\
\subfigure[A well-ripened banana.]{
\begin{minipage}[t!]{0.85\linewidth}
\centering
\includegraphics[width=\linewidth]{ 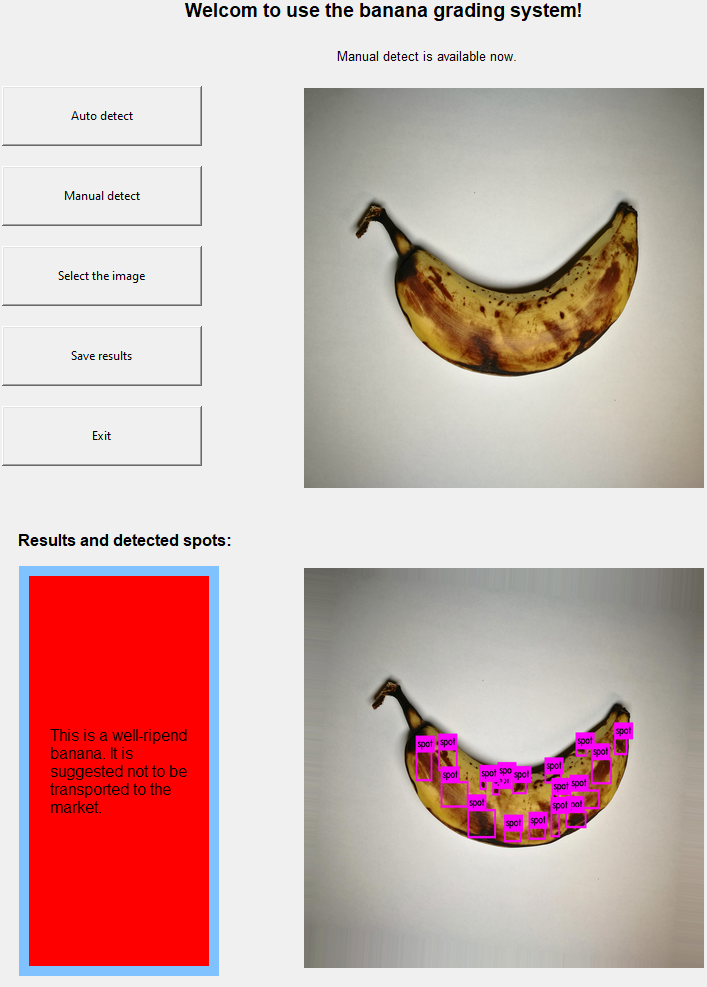}
\label{fig:app-rippened}
\end{minipage}%
}%
\centering
\caption{Examples of the analyzing results on the application.}
\end{figure}

The input (the upper image) and output (the lower image) of a sample banana is shown in Fig.~\ref{fig:app-unrippened}. The output image is the same as the input image as this banana is so unripened that there is no defective area on the peel. As a result, the context box suggested that this banana be transported to the market, and the green background means it passed the evaluation. When it comes to a Machine Vision System, this banana will be sent to a conveyor belt, which is leading to the good condition banana group after its quality is determined.

Compared to the unripened banana, Fig.~\ref{fig:app-rippened} illustrates an over-ripened sample banana. In this case, the context box is red to alarm the user that this banana is not good enough to be sent to the market. Then this banana will be directed to the defective banana area.

\section{Conclusion} \label{sec7}
This paper proposed a food grading system and focused on the novel two-layer classifier, which is the core of the system and can grade bananas according to their ripeness levels. Because there is redundant information in the original images, and this information will decrease the classification accuracy, a feature vector composed of the essential color and texture information was formed. The experiment results illustrate that the SVM classifier excels KNN, Random Forest, and Naive Bayes. The extracted feature vector assisted the SVM classifier to achieve an accuracy of 98.5\%. Then the YOLOv3 system continued to detect the ripened bananas' defective areas and separated them into the mid-ripened and the well-ripened groups. This system accomplished the banana grading mission in terms of ripeness and conquered the difficulty of detecting and outputting small defected areas on bananas. The design of the classification layers' distribution combines edge computing and cloud computing, thereby alleviating the pressure of network communication, saving computing resources, and shortening the feedback delay. This system is also applicable for grading and locating the defective areas of other varieties of fruit that have a significant change in color during ripening, such as apples, mangoes, and tomatoes.

The future work will continue to improve the second-layer classifier's performance by labelling the defected areas more precisely and fine-tuning the model, and improving the application to a smartphone version so that the banana evaluation could be realized with more ease.

\bibliography{V3revisedPetros.bib}

\end{document}